\documentclass[10pt,twocolumn,letterpaper]{article}

\usepackage[pagenumbers]{cvpr} 
\makeatletter
\@namedef{ver@everyshi.sty}{}
\makeatother
\usepackage{tikz}

\usepackage{multirow}
\newtheorem{definition}{Definition}

\newcounter{example}

\usepackage[ruled,vlined,linesnumbered]{algorithm2e}
\usepackage{tcolorbox}
\usepackage{tabularx}
\usepackage{graphicx}
\usepackage{amsmath}
\usepackage{amssymb}
\usepackage{mathrsfs}
\usepackage{booktabs}
\usepackage{xcolor}
\usepackage{colortbl}
\usepackage{bbm}
\usepackage{changepage}
\usepackage{balance}
\usepackage{amsmath}
\allowdisplaybreaks[4]
\definecolor{atomictangerine}{rgb}{1.0, 0.6, 0.4}
\definecolor{apricot}{rgb}{0.98, 0.81, 0.69}
\definecolor{antiquewhite}{rgb}{0.98, 0.92, 0.84}
\definecolor{Gray}{gray}{0.40}
\definecolor{LightCyan}{rgb}{0.88,1,1}
\definecolor{ashgrey}{rgb}{0.7, 0.75, 0.71}
\definecolor{darkgray}{gray}{0.80}
\definecolor{lightorange}{rgb}{0.98, 0.75, 0.56}
\definecolor{lo1}{rgb}{0.98, 0.93, 0.84}
\definecolor{lo2}{rgb}{0.98, 0.81, 0.69}
\definecolor{lo3}{rgb}{0.98, 0.60, 0.41}
\newcolumntype{L}[1]{>{\raggedright\let\newline\\\arraybackslash\hspace{0pt}}m{#1}}
\newcolumntype{C}[1]{>{\centering\let\newline\\\arraybackslash\hspace{0pt}}m{#1}}
\newcolumntype{R}[1]{>{\raggedleft\let\newline\\\arraybackslash\hspace{0pt}}m{#1}}

\usepackage[capitalize]{cleveref}
\crefname{section}{Sec.}{Secs.}
\Crefname{section}{Section}{Sections}
\Crefname{table}{Table}{Tables}
\crefname{table}{Tab.}{Tabs.}

\def\cvprPaperID{927} 
\def\confName{CVPR}
\def\confYear{2023}

\SetCommentSty{mycommfont}

\begin{document}


\title{
Boosting Verified Training for Robust Image Classifications via Abstraction}

\author{Zhaodi Zhang $^{1,2,4}$, Zhiyi Xue $^2$, Yang Chen $^2$, Si Liu $^3$, Yueling Zhang $^2$, Jing Liu $^1$, Min Zhang $^2$\\
 $^1$ Shanghai Key Laboratory of Trustworthy Computing \\
 $^2$ East China Normal University \\
 $^3$ ETH Z{\"u}rich $^4$ Chengdu Education Research Institute
}

\maketitle



\begin{abstract}
\vspace{-1mm}
This paper proposes a novel, abstraction-based, certified training method for robust image classifiers. Via abstraction, all perturbed images are mapped  into intervals before feeding into neural networks for training.
By training on intervals, all the perturbed images that are mapped to the same interval  are classified as the same label, rendering the variance of training sets to be small and the loss landscape of the models to be smooth. 
Consequently, our approach significantly improves the robustness of trained models. 
For the abstraction, 
our training method also enables a  sound and complete black-box verification approach, which is orthogonal and scalable to arbitrary types of neural networks regardless of their sizes and architectures. We evaluate our method on a wide range of benchmarks in different scales. The experimental results show that our method outperforms state of the art by (i) reducing the verified errors of trained models up to 95.64\%; (ii) totally achieving up to 602.50x speedup; and (iii) scaling up to larger models with up to 138 million trainable parameters. The demo is available at \url{ https://github.com/zhangzhaodi233/ABSCERT.git}.
\end{abstract}
\vspace{-3mm}

\section{Introduction}

The robustness of image classifications based on neural networks is attracting more attention than ever due to  their applications to safety-critical domains such as self-driving \cite{Wu_2017_CVPR_Workshops} and medical diagnosis \cite{titano2018automated}. There has been a considerable amount of work on training robust neural networks  against adversarial perturbations \cite{DBLP:conf/icml/AthalyeC018, DBLP:conf/ccs/Carlini017, DBLP:conf/sp/Carlini017, DBLP:journals/corr/GoodfellowSS14, DBLP:conf/iclr/MadryMSTV18, DBLP:conf/sp/PapernotM0JS16, DBLP:conf/cvpr/XiaoYLDL19, DBLP:conf/ijcai/XiaoLZHLS18, DBLP:conf/iclr/XiaoZ0HLS18, DBLP:conf/cvpr/EykholtEF0RXPKS18, DBLP:conf/acl/HsiehYCZC18, DBLP:conf/iclr/XuLZCZFEWL19, DBLP:conf/iclr/ZhangCSBDH19}. Conventional defending approaches  augment the training set with  adversarial examples \cite{DBLP:conf/sp/PapernotM0JS16, DBLP:conf/iclr/GuoRCM18, DBLP:conf/iclr/SongKNEK18, DBLP:conf/iclr/BuckmanRRG18, DBLP:conf/iclr/Ma0WEWSSHB18, DBLP:conf/iclr/SamangoueiKC18, DBLP:conf/eccv/XiaoDLYLS18, DBLP:conf/iccv/XiaoDLLEYSLM19}. 
They target only specific adversaries, depending on how the adversarial samples are generated \cite{DBLP:conf/sp/Carlini017},  but cannot provide robustness guarantees   \cite{DBLP:conf/iclr/BalunovicV20, DBLP:conf/aaai/FanL21, DBLP:conf/cvpr/LyuGWXZL21}. 

Recent approaches attempt to train certifiably robust models with guarantees 
\cite{DBLP:conf/iclr/BalunovicV20,  DBLP:conf/iccv/GowalDSBQUAMK19, DBLP:conf/icml/MirmanGV18, DBLP:conf/icml/WongK18, DBLP:conf/iclr/ZhangCXGSLBH20, DBLP:conf/cvpr/LyuGWXZL21,  DBLP:conf/aaai/FanL21, DBLP:conf/nips/LeeLPL21}. They rely on the robustness verification results of neural networks in the training process. Most of the existing verification approaches are based on symbolic interval propagation (SIP) \cite{DBLP:conf/iccv/GowalDSBQUAMK19,DBLP:conf/iccv/GowalDSBQUAMK19,DBLP:conf/cvpr/LyuGWXZL21,DBLP:conf/aaai/FanL21,DBLP:conf/nips/LeeLPL21}, by which intervals are \textit{symbolically input} into neural networks and propagated on a layer basis.
There are, however, mainly three obstacles for these approaches to be widely adopted:
(i) adding the verification results to the loss function for training brings limited improvement to the robustness of neural networks due to overestimation introduced in the verification phase;
(ii) they are time-consuming due to the high complexity of the verification problem \textit{per se}, e.g., NP-complete for the simplest ReLU-based fully connected feedforward neural networks \cite{DBLP:conf/cav/KatzBDJK17}; and
(iii) the verification is tied to specific types of neural networks in terms of  network architectures and  activation functions.

To overcome  the above obstacles, we propose a novel, abstraction-based approach for training verified robust image classifications whose inputs are 
\textit{numerical intervals}. 
Regarding (i), we abstract each pixel of an image into an interval before inputting it into the neural network. 
The interval is \textit{numerically} input to the neural network by assigning to two input neurons its lower and upper bounds, respectively. This guarantees that all the perturbations to the pixel in this interval do not alter the classification results, thereby improving the robustness of the network. Moreover, this imposes no overestimation in the 
training phase. To address the challenge (ii), we use forward propagation and back propagation only to train the network without extra time overhead.
Regarding (iii),  we treat the neural networks as black boxes since we  deal only with the input layer and do not change other layers. Hence, being agnostic to the actual neural network architectures, our approach can  scale up to fairly large neural networks. Additionally, 
we identify a crucial hyper-parameter, namely \emph{abstraction granularity}, 
which corresponds to the size of intervals used for training the networks. 
We propose a gradient descent-based algorithm to refine the abstraction granularity for training a more robust neural network. 

We  implement our method   in a tool  called \textsc{AbsCert} and assess it, together with existing approaches, on various benchmarks. The experimental results show that our approach 
reduces the verified errors of the trained neural networks up to 
95.64\%. Moreover, it totally achieves up to 
602.50x speedup.
Finally, it  can scale up to larger neural networks with up to 138 million trainable parameters and be applied to a wide range of neural networks.

\vspace{1ex}
\noindent \textbf{Contributions.} Overall, we provide (i) a
novel, 
abstraction-based training method for verified robust neural networks;
(ii) a companion black-box verification method for certifying the robustness of trained neural networks; 
(iii) a tool implementing our 
method; and
(iv) an extensive assessment of our 
method, together with existing approaches, on a wide range of benchmarks, which demonstrates our method's promising achievements.

\section{Robust Deep Neural Networks}\label{sec:prelim}



Huber~\cite{DBLP:books/wi/Huber81} introduces the concept of robustness in a broader sense: (\romannumeral1) the efficiency of the model should be reasonably good, and (\romannumeral2) the small disturbance applied to the input should only have a slight impact on the result of the model. 
The problem of robustness verification has been formally defined by Lyu \etal~\cite{DBLP:conf/cvpr/LyuGWXZL21}: 

\begin{definition}[Robustness verification]
Robustness verification aims to guarantee that a neural network outputs consistent classification results for all inputs in a set which is usually represented as an $l_p$ norm ball around the clean image $x$: $\mathbbm{B}(x, \epsilon) = \{x' |\left \Vert x' - x \Vert_p \leq \epsilon \right \}$. 
\end{definition}

\begin{figure}
 \begin{subfigure}{0.49\linewidth}
	\centering
	\includegraphics[width=0.9\linewidth]{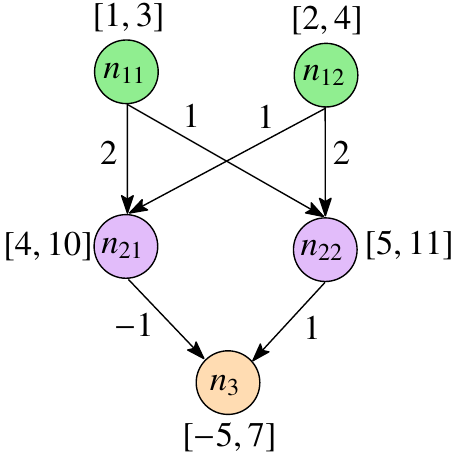}
	\caption{IBP}
\end{subfigure}
\centering
\begin{subfigure}{0.49\linewidth}
	\includegraphics[width=0.9\linewidth]{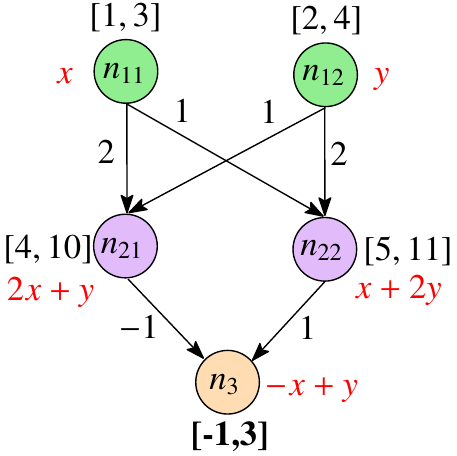}
	\caption{SIP}
\end{subfigure}
\vspace{-2mm}
\caption{IBP and SIP \cite{DBLP:conf/uss/WangPWYJ18}.}
\vspace{-5mm}
\label{fig:ibp}
\end{figure}

The problem of neural network robustness verification has intrinsically high complexity  \cite{DBLP:journals/corr/abs-2208-09872}. Many approaches rely on symbolic interval bound propagation to 
simplify the problem at the price of sacrificing the completeness 
 \cite{DBLP:journals/corr/abs-2208-09872,DBLP:conf/cav/JinTZWZ22,DBLP:conf/cvpr/LyuGWXZL21,DBLP:conf/iclr/ZhangCXGSLBH20}. Figure \ref{fig:ibp} illustratively compares the difference of interval bound propagation (IBP) and SIP. 
 Intuitively, when we know the domain of each input, we can estimate the output range by propagating the inputs symbolically throughout the neural network. According to the output range, we can prove/disprove the robustness of the neural network. See \cite{DBLP:conf/uss/WangPWYJ18} for more details.

To accelerate the propagation, the non-linear activation functions need to over-approximate using linear constraints, therefore rendering the output range overestimated. 
More hidden layers in a network imply a larger overestimation because the overestimation is accumulated layer by layer. A large overestimation easily causes failure to verification. Therefore, many efforts are being made to tighten the over-approximation \cite{zhang2018efficient,boopathy2019cnn,wu2021tightening}. Unfortunately, it has been proved there is a theoretical barrier \cite{salman2019convex,DBLP:journals/corr/abs-2208-09872,tjandraatmadja2020convex}.

\section{The Abstraction-Based Training  Approach}\label{sec:method}

In this section, we present our abstraction-based method for training neural networks. We define the training problem in Section \ref{3.1}. Section \ref{3.2} and Section \ref{3.3} give the abstraction procedure and the training procedure. 
Throughout these two sections, we also illustrate how varying
network inputs can affect the loss function,  which in turn contributes to the robustness of trained models.

\subsection{Formulating the Training Problem}
\label{3.1}

We solve the following training problem:

\begin{center}
	\begin{tcolorbox}[colback=gray!10,
		colframe=blue!40!white,
		boxrule=0.5pt,
		]
		\emph{Given a training set and a testing set of images and a perturbation distance $\epsilon$, our goal is to train an image classifier $f$ that is provably robust on $\mathbb{B}(x,\epsilon)$ for each image $x$ in the set while $f$ has a low verified error on the testing set. }
	\end{tcolorbox}
\end{center}

As the target image classifier $f$ must be guaranteed robust on $\mathbb{B}(x,\epsilon)$, $f$ returns the same classification results for all the perturbed images in $\mathbb{B}(x,\epsilon)$. Our idea is to (i) map $\mathbb{B}(x,\epsilon)$ to an interval vector $(I_1,I_2,\ldots I_n)$ by an abstraction function $\phi$ such that $\phi(\mathbb{B}(x,\epsilon))=(I_1,I_2,\ldots I_n)$, and (ii) train an image classifier $f'$ which takes as input the interval vector  and returns a classification result as $x$ is labeled, i.e., $\arg\max_{y\in Y}f'(I_1,I_2,\ldots,I_n)=y_{true}$ with $y_{true}$ the ground truth label of $x$.  The target image classifier $f$ is a composition of $f'$ and $\phi$. Apparently, $f$ is provably robust on  $\mathbb{B}(x,\epsilon)$ since all the images in $\mathbb{B}(x,\epsilon)$ are mapped to $(I_1,I_2,\ldots,I_n)$ which is classified to $y_{true}$ by $f'$.

\begin{figure}
	\centering
	\includegraphics[width=0.48\textwidth]{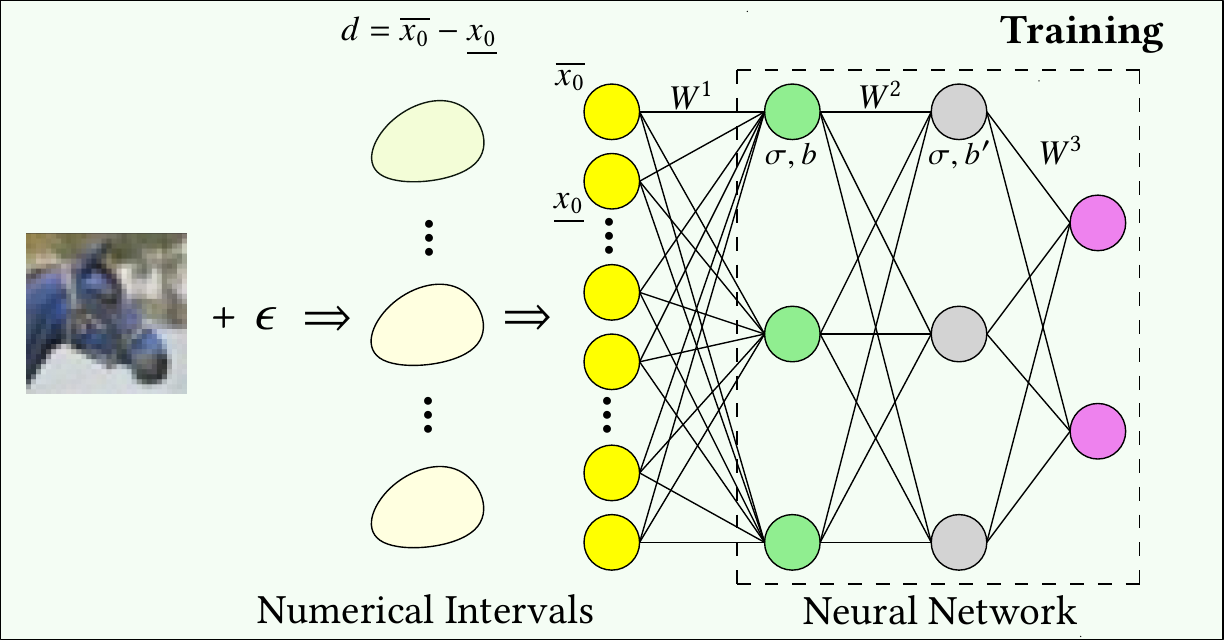}
	\caption{Training on numerical intervals. 
	}
	\label{training}
 \vspace{-1mm}
\end{figure}

To feed the training intervals to the neural network, the number of neurons in the input layer is doubled. As shown in Figure \ref{training}, the upper bound and lower bound of each training interval are input to these neurons. 
Namely, the neurons in the input layer do not correspond to a pixel but the upper and lower bounds of each training interval. Any perturbation interval mapped to the same training interval will be fed into the neural network with the same upper and lower bounds. Other training parameters and settings are the same as the traditional training process.

\subsection{Abstraction of Perturbed Images}
\label{3.2}
\subsubsection{From Perturbed Interval to Training Interval}

We first introduce the abstraction function $\phi:\mathcal{I}^n\rightarrow \mathbb{I}^n$, where 
$\mathcal{I}^n$ is the set of all interval vectors,  
$\mathbb{I}^n$ is a finite set of interval vectors on which neural networks are trained, and $n$ denotes the size of the interval vectors. The 
norm ball $\mathbb{B}(x,\epsilon)$ of
$x$ under $\epsilon$ is essentially an interval vector in $\mathcal{I}^n$. We call the elements in $\mathcal{I}^n$ \textit{perturbation interval vectors}. 
 
Let $[-1, 1]$ be the range of the complete input space. We divide $[-1, 1]$ evenly into sub-intervals of size $d_i$ and denote the set of all the sub-intervals as $\mathbb{I}_i$. Then, we obtain $\mathbb{I}^n$ as the Cartesian product of all $\mathbb{I}_i$, i.e., $\Pi^{n}_{i=1} \mathbb{I}_i$. 
Because we train image classifiers on $\mathbb{I}^n$, we call the elements in $\mathbb{I}^n$ \textit{training interval vectors}. 
We use $\overline{d}$ to denote the integer vector $(d_1,d_2,\ldots,d_n)$ and call it \emph{abstraction granularity}. In our abstraction process, all values in $\overline{d}$ are the same, so we use a constant $d$ to represent it. 

Because perturbation interval vectors are infinite, it is impossible to enumerate all perturbation interval vectors for training. We abstract and map them onto a finite number of training interval vectors in $\mathbb{I}^n$. The purpose of the abstraction is to ease the follow-up robustness verification by restricting the infinite perturbation space to finite training space. In addition, the abstraction function is an element-wise operation. In such cases, we introduce our mapping function with an interval as an example. 

In order to ensure that the perturbation interval can only be mapped to a unique training interval, we make the following constraints on the mapping process: We first restrict the abstraction granularity $d$ to be greater than or equal to twice the perturbation size $\epsilon$, for the purpose of guaranteeing that a perturbation interval has an intersection with at most two training intervals. We divide the mapping into three cases: (1) If a training interval contains the perturbation interval, the perturbation interval will be mapped to the training interval; (2) If the perturbation interval has an intersection with two training intervals unequally, the perturbation interval will be mapped to the training interval with the larger coverage area; (3) If the perturbation interval has an intersection with two training intervals equally, the perturbation interval will be mapped to the training interval with a larger value. In this way, we map the perturbation interval on the unique training interval, and then we can train the neural network on these training intervals.

\setlength{\textfloatsep}{10pt}
\begin{algorithm}[t]
	\renewcommand{\thealgocf}{1}
	\SetKwData{Left}{left}\SetKwData{This}{this}\SetKwData{Up}{up}
	\SetKwFunction{Union}{Union}\SetKwFunction{FindCompress}{FindCompress}
	\SetKwInOut{Input}{Input}\SetKwInOut{Output}{Output}
	\caption{ Abstraction of Perturbations $\phi$.}
	\label{abstraction_alg}
	\Input{~$\mathcal{I}$: perturbation interval vector;\\
		$d$: abstraction granularity}
	\Output{~$\mathbb{I}$: training interval vector}
	Initialize $I_{init}', \mathbb{I}$\tcp*{ $I_{init}'$ is [-1,1], and $\mathbb{I}$ is an empty vector}
	$I'$ $\leftarrow$ Divide($I_{init}'$, $d$)\tcp*{Get sub-intervals with size d}
		\For{i in len($\mathcal{I}$)}{ 
			\For{j in len($I'$)}{
				\If{min($\mathcal{I}_{i}$) $>$ max($I_j'$)}{
					Continue \tcp*{No intersection}
				}
				\If{$\mathcal{I}_{i} \in I_j'$}{
					$\mathbb{I}_{i} = I_j'$\tcp*{$I_j'$ contains $\mathcal{I}_{i}$}
				}
				\Else{
					$len_r$ = max($\mathcal{I}_{i}$) - min($I_{j+1}'$); \\
					$len_l$ = max($I_j'$) - min($\mathcal{I}_{i}$);\\
					\If{$len_r\geq len_l$}{
						$\mathbb{I}_{i} = I_{j+1}'$\tcp*{Map to numerically larger}
					}
					\Else{
						$\mathbb{I}_{i} = I_j'$\tcp*{Map to numerically smaller}
					}
				}
			}
		}
	Return ${\mathbb{I}}$;
\end{algorithm}

Algorithm \ref{abstraction_alg} shows our abstraction function. Firstly, we initialize the complete input interval and an training interval vector (Line 1). Then,
we obtain the sub-intervals which each size is $d$ (Line 2).
If a training interval contains the perturbation interval, the perturbation interval will be mapped to the training interval (Lines 7-8). The perturbation interval at most intersects with two training intervals at the same time as $d \geq 2*\epsilon$. Let $len_r$ be the size of the intersection of the perturbation interval and the numerically larger training interval (Line 10). Let $len_l$ be the size of the intersection of the perturbation interval and the numerically smaller training interval (Line 11). If $len_r\geq len_l$, the perturbation interval will be mapped to the numerically larger training interval (Lines 12-13); otherwise, it will be mapped to the numerically smaller training interval (Lines 14-15).

\vspace{-2mm}
\subsubsection{Effect of Abstraction on Input}
\vspace{-2mm}
We now explain that the abstraction process results in a smaller variance of training intervals. 

In our training method, the value of each pixel is normalized to $[-1, 1]$. Considering the arbitrariness of the input value distribution, we assume that the input values are evenly distributed in $[-1, 1]$. We calculate the variance of input, and in such cases, the value we get is the maximum likelihood estimation of the actual value.

We calculate the variance of the input values with $d$ representing abstraction granularity and $\mathbb{I}$ representing the training intervals of an image. Note that in conventional methods, it is equivalent to $d = 0$, while in our method, $d$ is a positive number. In this way, for an input image, the variance of it after the abstraction process is:
\begin{align}
	\label{variance}
	D(\mathbb{I}) & = E((\mathbb{I})^2) - E(\mathbb{I})^2 \notag\\
	& = \underbrace{(-1 + \frac{d}{2})^2 + (-1 + \frac{3d}{2}^2) + \ldots + (1 - \frac{d}{2})^2}_{2/d \enspace items} \\
	& = \frac{4n^2-1}{6}. \notag
\end{align}
\noindent In Equation \ref{variance}, the variance of training interval is computed by the upper and lower bounds. For an abstraction-based trained neural network, a large abstraction granularity $d$ implies a small $n$, and consequently the variance of the training intervals becomes small. Apparently, intervals have a smaller variance than  concrete pixel values.

\begin{algorithm}[t]
    \renewcommand{\thealgocf}{2}
	\SetKwData{Left}{left}\SetKwData{This}{this}\SetKwData{Up}{up}
	\SetKwFunction{Union}{Union}\SetKwFunction{FindCompress}{FindCompress}
	\SetKwInOut{Input}{Input}\SetKwInOut{Output}{Output}
	\label{train_alg}
	\caption{\mbox{ Abstraction-based Training: \textsc{AbsTrain}}}
	\label{train_alg}
	\Input{~$\mathbf{X}$: training data; \\ ~$\mathbf{Y}$: ground-truth labels of training data; \\ ~$\epsilon$: perturbation radius; \\ ~$d$: abstraction granularity}
	\Output{~$f'$: a neural network, $\ell$: training loss}
	Initialize $f'$, $\ell$; \\ 
	\For{($X$, $Y$) in ($\bf X, \bf Y$)}{
    \For{($x$, $y$) in ($X$, $Y$)}{
		$\mathbb{I}$ $\leftarrow \phi(\mathbb{B}(x$, $\epsilon$), $d$) \tcp*{Get training intervals.} 
		$\ell \leftarrow \ell + $ $\mathcal{L}$($f'(\mathbb{I})$, $\mathbf{y}$) \tcp*{Accumulate the loss.}
	  }
        $f'\leftarrow$ Update($f'$, $\ell$) \tcp*{Update the parameters in $f'$.}
	}
	Return $f'$, $\ell$;
\end{algorithm}

\subsection{Training on Intervals}
\label{3.3}
\subsubsection{The Training Method}

When we get the training intervals, hyperparameter settings, such as the number of layers, the number of neurons in hidden layers, the loss function during training, etc., are all the same as the traditional training methods except for the number of neurons in the input layer.

Algorithm \ref{train_alg} shows the pseudo-code of the algorithm for training a neural network. The training dataset $\mathbf{X}$, the ground-truth labels $\mathbf{Y}$ corresponding to the dataset, the perturbation radius $\epsilon$, and the abstraction granularity $d$ are used as inputs. Firstly, a neural network is initialized. That is, the adjustable parameters of the neural network are initialized randomly (Line 1). 
For each image of training dataset $\mathbf{X}$,
the perturbation intervals are mapped to training intervals (Line 4). According to the current adjustable parameters, 
the cross-entropy loss is calculated (Line 5). Finally, the backward propagation is performed according to the cross-entropy loss, and the values of the adjustable parameters are updated (Line 6). A trained neural network is returned when the loss no longer decreases.

\vspace{-2mm}
\subsubsection{Smoothing Loss Landscape}

We illustrate that the small variance of input results in a smooth loss landscape during training. Loss landscape is the characterization of loss functions. For example, smooth loss landscape means that the size of 
the connected region around the minimum where the training loss remains low, showing convex contours in the center, 
while sharp loss landscape shows not convex but chaotic to that region~\cite{DBLP:conf/nips/Li0TSG18}. 

We investigate the smoothness of the loss landscape from both theoretical and experimental perspectives. 
For a classification problem, the loss function is usually cross-entropy loss. We use $y$ to represent 
the ground-truth label, and $\hat{y}$ to represent the prediction of the neural network. In such cases, we explore the relationship 
between $\mathbb{I}$ and loss function. 
\begin{equation}
	\begin{aligned}
		\mathcal{L}(y, \hat{y}) 
		& = CrossEntropyLoss(y, \hat{y}) \\
		& = \sum_{i=1}^{c}y_i \cdot (-log(\hat{y_i})) \\
		& = -log(max(A \cdot \mathbb{I} + b))
		\label{eq:lossfunction}
	\end{aligned}
\end{equation}
where $c$ represents the number of 
classes, $y_i$ is the one-hot encoding of labels, 
$y_{true}$ is the output value corresponding to the correct label, and $A$ and $b$ are the parameters of the neural network.
For parameter space, a smooth loss landscape means that when the value of trainable parameters of the neural network gradually deviates 
from the optimal value, the loss increases slowly with it. In other words, the first-order partial derivative of 
Equation \ref{eq:lossfunction} with respect to $A$ and $b$ should be a constant or a value with less variation. 
The two partial derivatives are:
\begin{equation}
	\begin{aligned}
		\frac{\partial \mathcal{L}(y, \hat{y})}{\partial A} 
		& = \frac{\partial (-log(max(A \cdot \mathbb{I} + b)))}{\partial A} \\
		& = -\frac{\mathbb{I}}{max(A \cdot \mathbb{I} + b)}
		\label{eq:derivative_A}
	\end{aligned}
\end{equation}
\begin{equation}
	\begin{aligned}
		\frac{\partial \mathcal{L}(y, \hat{y})}{\partial b} 
		& = \frac{\partial (-log(max(A \cdot \mathbb{I} + b)))}{\partial b} \\
		& = -\frac{1}{max(A \cdot \mathbb{I} + b)}
		\label{eq:derivative_b}
	\end{aligned}
\end{equation}

With fixed $A$ and $b$, we discuss the value of Equation \ref{eq:derivative_A} for different training intervals.
If $b = 0$, we get a constant $-\frac{1}{A}$. Obviously, the loss landscape is smooth in this case. If $b \neq 0$, we write the derivative as $-\frac{1}{A + b \cdot (\mathbb{I})^{-1}}$. In our method, the variance of $\mathbb{I}$ is small. 
That is, the value of $\mathbb{I}$ concentrates around a fixed value. Thus, the value of the derivative varies in a small range, and the loss landscape is smoother. For Equation \ref{eq:derivative_b}, $max(A \cdot \mathbb{I} + b)$ is large as it is the value corresponding to the ground-truth label.
Hence, Equation \ref{eq:derivative_b} is close to $0$, and the loss landscape is smooth. 

\begin{figure}
	\centering
	\begin{subfigure}{0.49\linewidth}
		\centering
		\includegraphics[width=\linewidth]{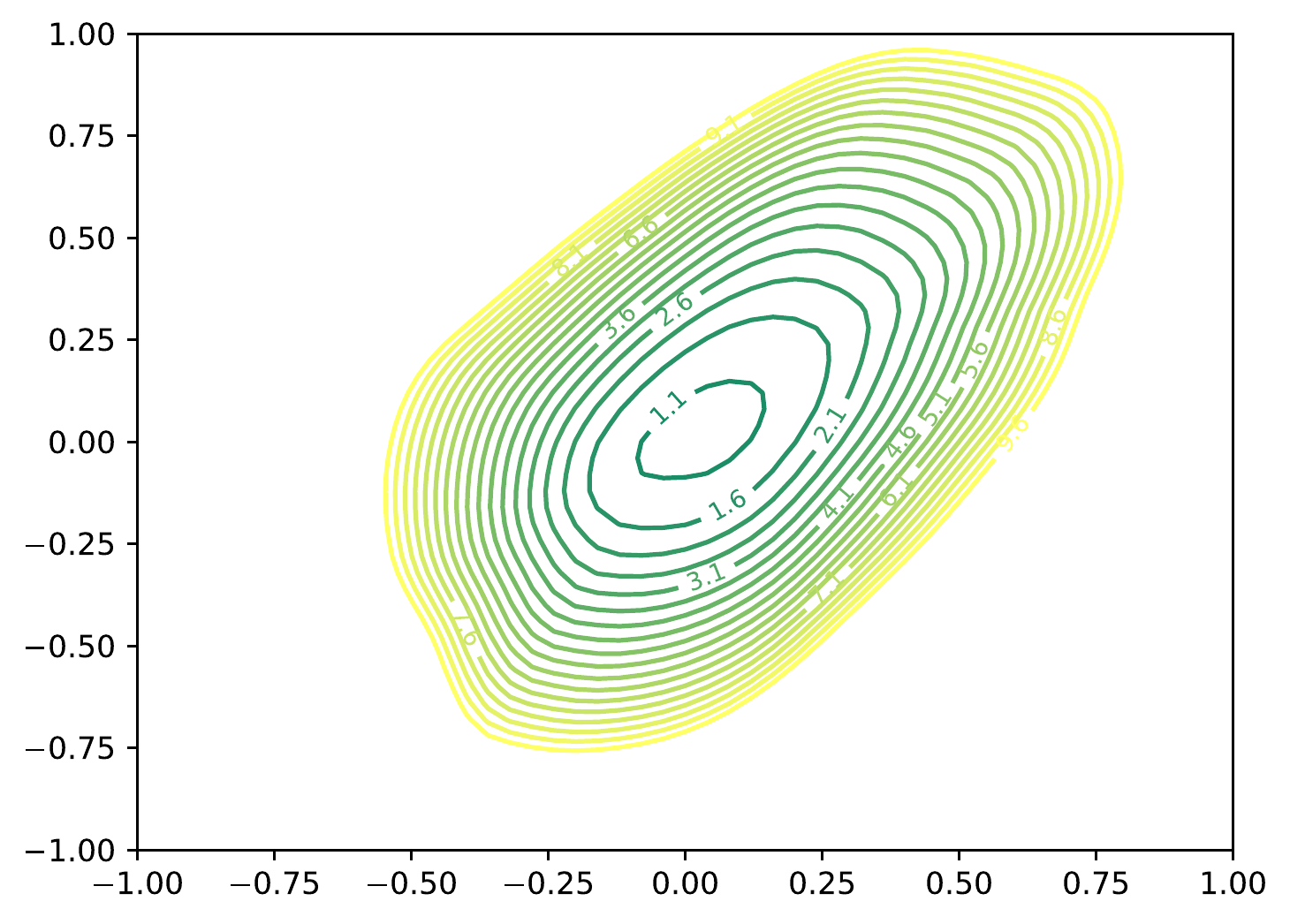}
		\caption{$d = 0$}
	\end{subfigure}
	\centering
	\begin{subfigure}{0.49\linewidth}
		\includegraphics[width=\linewidth]{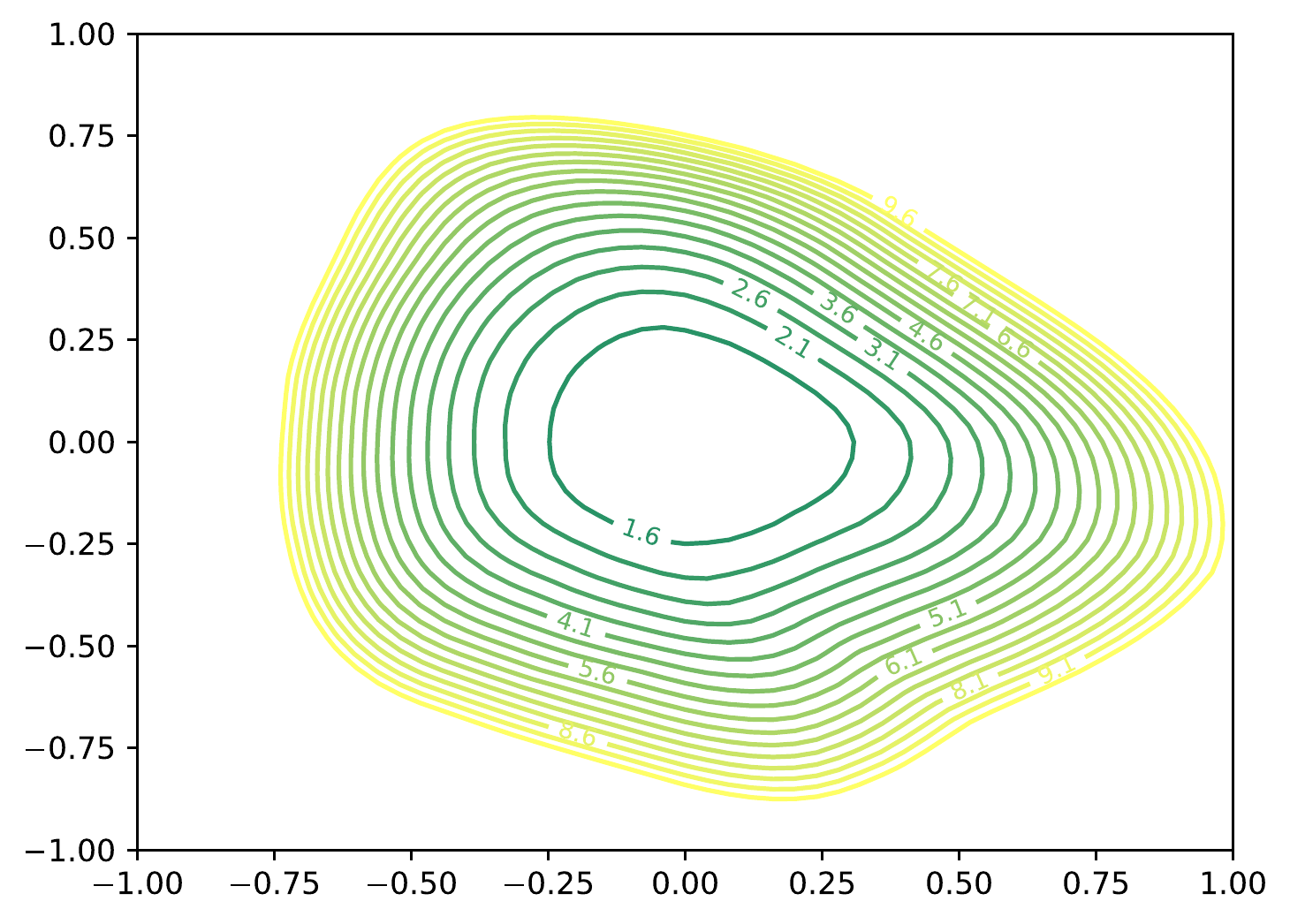}
		\caption{$d = 0.025$}
	\end{subfigure}
	\caption{Visualization of DM-small's loss landscape ~\cite{DBLP:conf/iclr/ZhangCXGSLBH20} trained when $d = 0$ and $d = 0.025$. 
	}
	\label{fig:loss_landscape}
\end{figure}

In summary, the small variance of $\mathbb{I}$ results in a smooth loss landscape. 
As an example, we utilize the tool in \cite{DBLP:conf/nips/Li0TSG18} to visualize in Figure \ref{fig:loss_landscape} the loss landscape of two neural networks that are trained with $d = 0$ and $d = 0.025$, respectively. 
Figure \ref{fig:loss_landscape} depicts that the neural network trained with $d = 0.025$ has a broader loss landscape than the one trained with $d=0$, and the loss increases more smoothly in every direction.

\vspace{-3mm}
\subsubsection{The Robustness of Trained Neural Networks}

Training on numerical intervals guarantees that all the perturbed intervals that are mapped to the same numerical interval will have the same classification result. Intuitively, if a numerical interval represents more concrete values, even if the concrete value is slightly disturbed, these disturbed values will be mapped to the same numerical interval with a high probability. Consequently, the classification of the numerical interval according to the neural network still remains unchanged. In this sense, we say that the perturbation is \textit{dissolved} by the abstraction, and therefore the neural network's robustness is improved. 

Theoretically, we show that the smooth loss landscape due to the abstraction in  training contributes to the robustness of neural networks. 
As shown in Figure \ref{fig:loss_landscape}, 
the loss of DM-small trained by \textsc{AbsCert} increases slowly and uniformly with the change of parameters, meaning that the growth rate of loss is slow in all directions of parameters' change. However, neural networks trained by conventional methods have a steep slope, which means that there is a direction where the loss increases rapidly. Obviously, smooth loss landscape is helpful for the optimization in training phase to find a global optimum. 

Moreover, in the process of our abstraction, an original pixel is first perturbed to a perturbed interval which is then mapped to a training interval for classification. This indicates that the loss is a constant because all the perturbed images of an image are mapped to a fixed set of training intervals, and the $\mathbb{I}^n$ in Equation \ref{eq:lossfunction} is never changed. 

Neural networks trained by our abstraction-based training method is more robust as the loss landscape is smooth in both parameter space and input space~\cite{DBLP:conf/ijcai/YuQLZWC19}. In particular, for parameter space, the model tends to find a global optimum in a reasonably efficient manner~\cite{DBLP:conf/nips/Li0TSG18, DBLP:conf/nips/WuX020}; for input space, the model is insensitive to the input perturbations ~\cite{DBLP:conf/cvpr/Moosavi-Dezfooli19}.

\section{Formal Verification and Granularity Tuning} 
\label{4}

\begin{algorithm}[t]
    \renewcommand{\thealgocf}{3}
	\SetKwData{Left}{left}\SetKwData{This}{this}\SetKwData{Up}{up}
	\SetKwFunction{Union}{Union}\SetKwFunction{FindCompress}{FindCompress}
	\SetKwInOut{Input}{Input}\SetKwInOut{Output}{Output}
	\caption{\mbox{Training with Granularity Tuning.}}
	\label{refine_alg}
	\Input{~$\mathbf{X}$: training data; ~$\mathbf{X_{test}}$: test data;\\ ~$\mathbf{Y}$: ground-truth labels of training data;\\ ~$\mathbf{Y_{test}}$: ground-truth labels of test data;\\ ~$\epsilon$: perturbation radius;\\ ~$d_{step}$: step size of abstraction granularity}
	\Output{$f_{out}$: verified robust neural network;\\
	$d_{out}$: the best abstraction granularity}
	Initialize $d,error,e, f_{out}, d_{out}$\tcp*{$e:$ verified error.}
	\While{$d\geq 2*\epsilon$}{
	    $(f,\ell)\leftarrow$  \textsc{AbsTrain}($\mathbf{X}$, $\mathbf{Y}$, $\epsilon$, $d$)\tcp*{$\ell$: training loss.} 
	    
	    $e\leftarrow $ \textsc{Verify}($f$, $\mathbf{X_{test}}$,  $\mathbf{Y_{test}}$, $\epsilon$, $d$)\tcp*{Verification.} 
	    
	    \If{$e < error$}{
	        $error\leftarrow e$; \\
	        $f_{out}\leftarrow f$;\\
	        $d_{out}\leftarrow d$; 
	    }
	    
	    $\mathbb{I}$ $\leftarrow \Phi(\mathbf{X}$, $\epsilon$, $d$) \tcp*{Map to training interval.} 
	    
	    $\overline{G}, \underline{G}\leftarrow\ell'(\overline{\mathbb{I}}),\ell'(\underline{\mathbb{I}}) $\tcp*{Obtain bounds' gradient,}
	    
	    
	    \If{$\overline{G}\leq 0 \wedge \underline{G}\geq 0$}{
	        Break\tcp*{Stop when G guides d to increase} 
	    }
	    \Else{
	        $d \leftarrow  d - d_{step}$\tcp* {Decrease d and continue}
	    }
	}
	Return $f_{out}, d_{out}$;
\end{algorithm}

In this section we introduce a verification-based method for 
tuning the abstraction granularity to train 
robust models. Due to the finiteness of $\mathbb{I}^n$, verification procedure can be conducted in a black-box manner, which is both sound and complete. Based on verification results, we can 
tune the abstraction granularity to obtain finer training intervals.

\subsection{Black-box Robustness Verification}

We propose a black-box verification method $\textsc{Verify}(\cdot)$ for neural networks trained by our approach. Given a neural network $f$, a test set $\mathbf{X_{test}}, \mathbf{Y_{test}}$, a perturbation distance $\epsilon$ and an abstraction granularity  $d$, $\textsc{Verify}(\cdot)$ returns the verified error on the set. 
The verification procedure is straightforward. First, for each $x\in \mathbf{X_{test}}$, we compute the set $\mathbb{I}$ of training interval vectors of $\mathbb{B}(x,\epsilon)$ using the same abstraction function $\phi$ in  Algorithm \ref{abstraction_alg}.  
Then, we feed each interval vector in $\mathbb{I}$ into $f$ and check if the classification result is consistent with the ground-truth label of $x$. The verified error $e$ is the ratio of the inconsistent cases in the test set. 

Our verification method is both sound and complete due to the finiteness of $\mathbb{I}$: 
$f$ is robust on $\mathbb{B}(x,\epsilon)$ if and only if $f$ returns the same label on all the interval vectors in   $\mathbb{I}$ as the one of $x$. Another advantage 
is that it treats $f$ as a black box. Therefore, our verification method is orthogonal and scalable to arbitrary models.  

\subsection{Tuning Abstraction Granularity} \label{3.4}

When the verified error of a trained neural network is large, we can reduce it by tuning the abstraction granularity and re-train the model on the refined training set of interval vectors. We propose a gradient descent-based 
algorithm to explore the best $d$ in the abstraction space. 

Algorithm \ref{refine_alg} shows the tuning and re-training process. First, our algorithm initializes abstraction granularity $d$ and verified error $error$ (Line 1). The tuning process repeats until $d < 2\epsilon$, which means that the size of a training interval is not less than that of the perturbed interval (Lines 2-15). Then a neural network and training loss toward training intervals with $d$ are obtained by calling the Abstraction-based Training function (Algorithm \ref{train_alg}) (Line 3). Get the neural network's verified errors on the test dataset (Line 4). If the neural network's verified errors are smaller, we save the neural network and abstraction granularity (Lines 5-8). Next, training intervals are obtained (Line 9). Then the upper and lower bounds' gradient of the training intervals are obtained (Lines 10). 
If $\overline{G} \leq 0$ and $\underline{G} \geq 0$, the algorithm will terminate (Lines 11-12); otherwise, abstraction granularity will be updated to a smaller granularity (Lines 13-14). 
When the algorithm terminates, we obtain the neural network with the lowest verified error.

\renewcommand\arraystretch{1.5}
\section{Experiment}\label{sec:expri}

\begin{table*}[ht!]\Huge
    \centering
    \caption{Verified errors (\%) of models trained by LossLandscapeMatters (LLM), LBP\&Ramp (LBP), AdvIBP (AIBP), and \textsc{AbsCert} (AC) on MNIST and CIFAR-10 datasets. 
    ``--'' means that the publicly available code of LBP does not support the CIFAR-10 dataset.}
    \vspace{-2mm}
    \resizebox{\textwidth}{23mm} {
    \begin{tabular}{c|c|c|c|c|c|c|c|c|c|c|c|c|c|c|c|c|c|c|c|c|c|c}
        \hline
        \multirow{2}{*}{\textbf{Dataset}} &
        \multirow{2}{*}{\textbf{$\epsilon$}} &
        \multicolumn{7}{c|}{\textbf{DM-small}} &
        \multicolumn{7}{c|}{\textbf{DM-medium}} &
        \multicolumn{7}{c}{\textbf{DM-large}} \\
        \cline{3-23}
         & & \makebox[0.07\textwidth][c]{AC} & 
         \makebox[0.07\textwidth][c]{LLM} & 
         \makebox[0.07\textwidth][c]{Impr.(\%)} & 
         \makebox[0.07\textwidth][c]{LBP} & 
         \makebox[0.07\textwidth][c]{Impr.(\%)} & 
         \makebox[0.07\textwidth][c]{AIBP} & 
         \makebox[0.07\textwidth][c]{Impr.(\%)} & 
         \makebox[0.07\textwidth][c]{AC} & 
         \makebox[0.07\textwidth][c]{LLM} & 
         \makebox[0.07\textwidth][c]{Impr.(\%)} & 
         \makebox[0.07\textwidth][c]{LBP} & 
         \makebox[0.07\textwidth][c]{Impr.(\%)} &
         \makebox[0.07\textwidth][c]{AIBP} &
         \makebox[0.07\textwidth][c]{Impr.(\%)} &
         \makebox[0.07\textwidth][c]{AC} &
         \makebox[0.07\textwidth][c]{LLM} &
         \makebox[0.07\textwidth][c]{Impr.(\%)} &
         \makebox[0.07\textwidth][c]{LBP} &
         \makebox[0.07\textwidth][c]{Impr.(\%)} &
         \makebox[0.07\textwidth][c]{AIBP} &
         \makebox[0.07\textwidth][c]{Impr.(\%)} \\
        \hline
        \multirow{4}{*}{MNIST}
        & 0.1 & 0.89 & 3.02 & \cellcolor{lo1} 70.53 $\uparrow$
                     & 4.09 & \cellcolor{lo1} 78.24 $\uparrow$
                     & 3.34 & \cellcolor{lo1} 73.35 $\uparrow$
              & 0.69 & 2.67 & \cellcolor{lo1} 74.16 $\uparrow$
                     & 3.47 & \cellcolor{lo2} 80.12 $\uparrow$
                     & 3.62 & \cellcolor{lo2} 80.94 $\uparrow$
              & 0.52 & 2.29 & \cellcolor{lo1} 77.29 $\uparrow$
                     & 2.93 & \cellcolor{lo2} 82.25 $\uparrow$
                     & 3.66 & \cellcolor{lo2} 85.79 $\uparrow$\\
        & 0.2 & 0.94 & 6.04 & \cellcolor{lo2} 84.44 $\uparrow$
                     & 5.54 & \cellcolor{lo2} 83.03 $\uparrow$
                     & 5.96 & \cellcolor{lo2} 84.23 $\uparrow$
              & 0.70 & 5.10 & \cellcolor{lo2} 86.27 $\uparrow$
                     & 4.73 & \cellcolor{lo2} 85.20 $\uparrow$
                     & 6.05 & \cellcolor{lo2} 88.43 $\uparrow$
              & 0.61 & 4.38 & \cellcolor{lo2} 86.07 $\uparrow$
                     & 3.96 & \cellcolor{lo2} 84.60 $\uparrow$
                     & 5.89 & \cellcolor{lo2} 89.64 $\uparrow$\\ 
        & 0.3 & 1.01 & 12.48 & \cellcolor{lo3} 91.91 $\uparrow$
                     & 8.11  & \cellcolor{lo2} 87.55 $\uparrow$
                     & 12.16 & \cellcolor{lo3} 91.69 $\uparrow$
              & 0.77 & 11.75 & \cellcolor{lo3} 93.45 $\uparrow$
                     & 7.02  & \cellcolor{lo2} 89.03 $\uparrow$
                     & 9.61  & \cellcolor{lo3} 91.99 $\uparrow$
              & 0.64 & 10.38 & \cellcolor{lo3} 93.83 $\uparrow$
                     & 6.14  & \cellcolor{lo2} 89.58 $\uparrow$
                     & 8.76  & \cellcolor{lo3} 92.69 $\uparrow$\\
        & 0.4 & 1.22 & 20.51 & \cellcolor{lo3} 94.05 $\uparrow$
                     & 13.03 & \cellcolor{lo3} 90.64 $\uparrow$
                     & 20.69 & \cellcolor{lo3} 94.10 $\uparrow$
              & 0.93 & 19.04 & \cellcolor{lo3} 95.12 $\uparrow$
                     & 11.59 & \cellcolor{lo3} 91.98 $\uparrow$
                     & 17.33 & \cellcolor{lo3} 94.63 $\uparrow$
              & 0.77 & 15.71 & \cellcolor{lo3} 95.10 $\uparrow$
                     & 10.48 & \cellcolor{lo3} 92.65 $\uparrow$
                     & 17.68 & \cellcolor{lo3} 95.64 $\uparrow$\\
        \hline
        \multirow{5}{*}{CIFAR-10}
        & $2/255$ & 25.52 & 50.95 & \cellcolor{lo1} 49.91 $\uparrow$
                                  & --    & -- 
                                  & 57.20 & \cellcolor{lo1} 55.38 $\uparrow$
                          & 16.40 & 49.83 & \cellcolor{lo1} 67.09 $\uparrow$
                                  & --     & -- 
                                  & 54.21 & \cellcolor{lo1} 69.75 $\uparrow$
                          & 13.81 & 48.20 & \cellcolor{lo1} 71.35 $\uparrow$
                                  & -- & --
                                  & 54.39 & \cellcolor{lo1} 74.61 $\uparrow$ \\
        & $4/255$ & 25.52 & 61.90 & \cellcolor{lo1} 58.77 $\uparrow$
                                  & --     & -- 
                                  & 65.30     & \cellcolor{lo1} 60.92 $\uparrow$
                          & 16.40 & 61.46 & \cellcolor{lo1} 73.32 $\uparrow$
                                  & --     & -- 
                                  & 62.63     & \cellcolor{lo1} 73.81 $\uparrow$ 
                          & 13.81 & 61.22 & \cellcolor{lo1} 77.44 $\uparrow$
                                  & --     & --
                                  & 61.95 & \cellcolor{lo1} 77.71 $\uparrow$ \\
        & $6/255$ & 25.52 & 68.36 & \cellcolor{lo1} 62.67 $\uparrow$
                                  & --    & --   
                                  & 70.20     & \cellcolor{lo1} 63.65 $\uparrow$
                          & 16.70 & 67.28 & \cellcolor{lo1} 75.18 $\uparrow$
                                  & --     & --
                                  & 67.69     & \cellcolor{lo1} 75.33 $\uparrow$ 
                          & 13.88 & 66.99 & \cellcolor{lo1} 79.28 $\uparrow$
                                  & --     & --
                                  & 67.56 & \cellcolor{lo1} 79.46 $\uparrow$ \\
        & $8/255$ & 25.52 & 71.92 & \cellcolor{lo1} 64.52 $\uparrow$
                                  & --     & -- 
                                  & 72.50 & \cellcolor{lo1} 64.80 $\uparrow$
                          & 16.93 & 70.54 & \cellcolor{lo1} 76.00 $\uparrow$
                                  & --    & --
                                  & 70.75 & \cellcolor{lo1} 76.07 $\uparrow$
                          & 13.88 & 70.35 & \cellcolor{lo2} 80.27 $\uparrow$
                                  & -- & --
                                  & 70.72 & \cellcolor{lo2} 80.37 $\uparrow$ \\
        & $16/255$ & 26.61 & 78.13 & \cellcolor{lo1} 65.94 $\uparrow$
                                   & --     & --  
                                   & 78.90 & \cellcolor{lo1} 66.27 $\uparrow$
                           & 17.16 & 78.27 & \cellcolor{lo1} 78.08 $\uparrow$
                                   & --     & --
                                   & 78.33 & \cellcolor{lo1} 78.09 $\uparrow$
                           & 14.12 & 78.03 & \cellcolor{lo2} 81.90 $\uparrow$
                                   & -- & -- 
                                   & 78.31 & \cellcolor{lo2} 81.97 $\uparrow$ \\
        \hline
        
    \end{tabular}
    }
    \label{tab:exp1}
    \vspace{-2mm}
\end{table*}

\begin{table}
    \centering
    \caption{Accuracy (\%) of models
    trained by LLM, LBP, AIBP, and \textsc{AC} (our method) on MNIST and CIFAR-10.
    }
    \vspace{-2mm}
    \setlength{\tabcolsep}{7.6pt}
    \scriptsize
    	\renewcommand{\arraystretch}{1.2}
    \begin{tabular}{c|c|c|c|c|c|c}
        \hline
          &  \textbf{Dataset} & \textbf{$\epsilon$} & AC & LLM & LBP & AIBP  \\
        \hline
        \multirow{9}{*}{\rotatebox{90}{DM-small}}  
		  & \multirow{4}{*}{MNIST} 
		    & 0.1 & \cellcolor{lo1} 99.11 & 98.43 & 96.63 & 98.36 \\
	      & & 0.2 & \cellcolor{lo1} 99.06 & 97.15 & 96.94 & 97.89  \\ 
	      & & 0.3 & \cellcolor{lo1} 98.99 & 94.38 & 96.65 & 96.35  \\
	      & & 0.4 & \cellcolor{lo1} 98.78 & 94.46 & 96.65 & 96.14 \\
		\cline{2-7}
	      & \multirow{5}{*}{CIFAR-10} 
		    & $2/255$ & \cellcolor{lo1} 74.48 & 64.70 & -- & 59.20   \\
	      & & $4/255$ & \cellcolor{lo1} 74.48 & 55.07 & -- & 49.69    \\
	      & & $6/255$ & \cellcolor{lo1} 74.48 & 49.29 & -- & 41.94    \\
	      & & $8/255$ & \cellcolor{lo1} 74.48 & 44.34 & -- & 39.52 \\
	      & & $16/255$ & \cellcolor{lo1} 73.39 & 32.58 & -- & 30.74 \\
	    \hline
		\multirow{9}{*}{\rotatebox{90}{DM-medium}}
		  & \multirow{4}{*}{MNIST}
            & 0.1 & \cellcolor{lo1} 99.31 & 98.76 & 97.37 & 98.65 \\
          & & 0.2 & \cellcolor{lo1} 99.30 & 98.13 & 97.36 & 98.42  \\ 
          & & 0.3 & \cellcolor{lo1} 99.23 & 95.04 & 97.35 & 97.45  \\
          & & 0.4 & \cellcolor{lo1} 99.07 & 93.60 & 97.36 & 97.44 \\
		\cline{2-7}
          & \multirow{5}{*}{CIFAR-10}
            & $2/255$ & \cellcolor{lo1} 83.60 & 66.00 & -- & 62.04   \\
          & & $4/255$ & \cellcolor{lo1} 83.60 & 55.09 & -- & 52.37    \\
          & & $6/255$ & \cellcolor{lo1} 83.30 & 48.38 & -- & 46.26    \\
          & & $8/255$ & \cellcolor{lo1} 83.07 & 41.19 & -- & 40.91 \\
          & & $16/255$ & \cellcolor{lo1} 82.84 & 33.65 & -- & 31.32 \\
        \hline
		\multirow{9}{*}{\rotatebox{90}{DM-large}}
		  & \multirow{4}{*}{MNIST}
            & 0.1 & \cellcolor{lo1} 99.48 & 98.95 & 97.79 & 98.96 \\
          & & 0.2 & \cellcolor{lo1} 99.39 & 98.41 & 97.79 & 98.47  \\ 
          & & 0.3 & \cellcolor{lo1} 99.36 & 95.90 & 97.77 & 98.05  \\
          & & 0.4 & \cellcolor{lo1} 99.23 & 96.14 & 97.79 & 97.78 \\
		\cline{2-7}
          & \multirow{5}{*}{CIFAR-10}
            & $2/255$ & \cellcolor{lo1} 86.19 & 62.50 & -- & 62.33   \\
          & & $4/255$ & \cellcolor{lo1} 86.19 & 56.99 & -- & 52.73    \\
          & & $6/255$ & \cellcolor{lo1} 86.12 & 49.58 & -- & 45.13    \\
          & & $8/255$ & \cellcolor{lo1} 86.12 & 43.26 & -- & 41.54 \\
          & & $16/255$ & \cellcolor{lo1} 85.88 & 33.03 & -- & 30.19 \\
        \hline
    \end{tabular}
    \label{tab:small}
\end{table}

We have implemented both our training and verification approaches in a tool called \textsc{AbsCert}. 
We evaluate \textsc{AbsCert}, together with existing approaches, on various public benchmarks with respect to both \emph{verified error} and \emph{training and verification time}.

By comparing with the state of the art, our goal is to demonstrate that
    \textsc{AbsCert} can train neural networks with lower verified errors (\textbf{Experiment \uppercase\expandafter{\romannumeral1}}),  incurs less computation overhead in both  training and verification (\textbf{Experiment \uppercase\expandafter{\romannumeral2}}), is applicable to a wide range of neural network architectures (\textbf{Experiment \uppercase\expandafter{\romannumeral3}}), and can scale up to larger models (\textbf{Experiment \uppercase\expandafter{\romannumeral4}}). 

\subsection{Benchmarks and Experimental Setup}
\noindent
\textbf{Competitors.} 
We consider three state-of-the-art provably robust training methods:
LossLandscapeMatters~\cite{DBLP:conf/nips/LeeLPL21}, LBP\&Ramp~\cite{DBLP:conf/cvpr/LyuGWXZL21}, and AdvIBP~\cite{DBLP:conf/aaai/FanL21}. 
All of them rely on linear approximation to train certifiably robust neural networks, which minimizes the upper bound on the worst-case loss against $l_\infty$ perturbations.  
We use their predefined optimal hyper-parameters  to train
neural networks on different perturbations $\epsilon$.

\begin{table*}[]\Large
    \centering
    \caption{Training time (s/epoch) and verification time (s) of LossLandscapeMatters (LLM), LBP\&Ramp (LBP), AdvIBP (AIBP), and \textsc{AbsCert} (AC). 
    ``--'' means that the publicly available code of  LBP does not support the CIFAR-10 dataset.}
    \vspace{-2mm}
    \resizebox{\textwidth}{!} {
        \begin{tabular}{c|l|c|c|c|c|c|c|c|c|c|c|c|c|c|c|c|c|c}
            \hline
            \multirow{2}{*}{\textbf{Dataset}} &
            \multirow{2}{*}{\hspace{6mm}\textbf{Model}} &
            \multicolumn{7}{c}{\textbf{Training Time}} & 
            \multicolumn{7}{|c}{\textbf{Verification Time}} & 
            \multicolumn{3}{|c}{\textbf{Total Speedup}}\\
            \cline{3-19}
            & & 
            \makebox[0.08\textwidth][c]{AC} &
            \makebox[0.08\textwidth][c]{LLM} &
            \makebox[0.08\textwidth][c]{SpeedUp} &
            \makebox[0.08\textwidth][c]{LBP} &
            \makebox[0.08\textwidth][c]{SpeedUp} &
            \makebox[0.08\textwidth][c]{AIBP} &
            \makebox[0.08\textwidth][c]{SpeedUp} &
            \makebox[0.08\textwidth][c]{AC} &
            \makebox[0.08\textwidth][c]{LLM} &
            \makebox[0.08\textwidth][c]{SpeedUp} &
            \makebox[0.08\textwidth][c]{LBP} &
            \makebox[0.08\textwidth][c]{SpeedUp} &
            \makebox[0.08\textwidth][c]{AIBP} &
            \makebox[0.08\textwidth][c]{SpeedUp} &
            \makebox[0.08\textwidth][c]{LLM} &
            \makebox[0.08\textwidth][c]{LBP} &
            \makebox[0.08\textwidth][c]{AIBP}\\
            \hline
            \multirow{3}{*}{MNIST} 
            & DM-small & 2.18 & 13.40 & \cellcolor{lo1} 6.14 $\uparrow$
                              & 6.54  & \cellcolor{lo1} 2.99 $\uparrow$
                              & 12.46 & \cellcolor{lo1} 5.72 $\uparrow$ 
                       & 0.54 & 5.77  & \cellcolor{lo2} 10.69 $\uparrow$
                              & 66.39 & \cellcolor{lo3} 122.94 $\uparrow$
                              & 2.35 & \cellcolor{lo1} 4.35 $\uparrow$ 
                              & \cellcolor{lo1} 7.05 $\uparrow$ & \cellcolor{lo2} 26.81 $\uparrow$ & \cellcolor{lo1} 5.44 $\uparrow$\\
            & DM-medium & 2.32 & 35.17 & \cellcolor{lo2} 15.18 $\uparrow$
                               & 11.99 & \cellcolor{lo1} 5.18  $\uparrow$
                               & 34.72 & \cellcolor{lo2} 14.97 $\uparrow$ 
                        & 0.56 & 6.19 & \cellcolor{lo2} 11.05 $\uparrow$ 
                               & 129.18  & \cellcolor{lo3} 230.68 $\uparrow$
                               & 2.63 & \cellcolor{lo1} 4.70 $\uparrow$ 
                               & \cellcolor{lo2} 14.36 $\uparrow$ & \cellcolor{lo2} 49.02 $\uparrow$ & \cellcolor{lo2} 12.97 $\uparrow$\\
            & DM-large & 4.28 & 113.14 & \cellcolor{lo2} 26.43 $\uparrow$
                              & 36.14  & \cellcolor{lo1} 8.44  $\uparrow$
                              & 99.89  & \cellcolor{lo2} 23.34 $\uparrow$ 
                       & 0.56 & 15.11 & \cellcolor{lo2} 26.98 $\uparrow$
                              & 337.40 & \cellcolor{lo3} 602.50 $\uparrow$
                              & 3.79 & \cellcolor{lo1} 6.77 $\uparrow$ 
                              & \cellcolor{lo2} 26.50 $\uparrow$ & \cellcolor{lo2} 77.18 $\uparrow$ & \cellcolor{lo2} 21.42 $\uparrow$\\
            \hline
            \multirow{3}{*}{CIFAR-10}
             & DM-small  & 3.31 & 15.00 & \cellcolor{lo1} 4.54 $\uparrow$
                         & -- & -- 
                         & 13.76 & \cellcolor{lo1} 4.16 $\uparrow$
                         & 0.64 & 7.59 & \cellcolor{lo2} 11.86 $\uparrow$
                         & -- & -- 
                         & 3.53 & \cellcolor{lo1} 5.52 $\uparrow$
                         & \cellcolor{lo1} 5.72 $\uparrow$ & -- & \cellcolor{lo1} 4.38 $\uparrow$\\
             & DM-medium & 3.58 & 31.40 & \cellcolor{lo1} 8.77 $\uparrow$ 
                         & -- & --
                         & 35.84 & \cellcolor{lo2} 10.01 $\uparrow$
                         & 0.66 & 7.68 & \cellcolor{lo2} 11.64 $\uparrow$ 
                         & -- & -- 
                         & 3.99 & \cellcolor{lo1} 6.04 $\uparrow$ 
                         & \cellcolor{lo1} 9.22 $\uparrow$ & -- & \cellcolor{lo1} 9.39 $\uparrow$\\
             & DM-large  & 5.29 & 123.22 & \cellcolor{lo2} 23.29 $\uparrow$ 
                         & -- & --
                         & 163.31 & \cellcolor{lo2} 30.87 $\uparrow$ 
                         & 0.76 & 18.45 & \cellcolor{lo2} 24.28 $\uparrow$
                         & -- & -- 
                         & 6.60 & \cellcolor{lo1} 8.68 $\uparrow$ 
                         & \cellcolor{lo2} 23.42 $\uparrow$ & -- & \cellcolor{lo2} 28.08 $\uparrow$\\
             \hline
        \end{tabular}
    }
    \vspace{-2mm}
    \label{tab:exp2}
\end{table*}

\vspace{1ex}
\noindent
\textbf{Datasets and Networks.}
We conduct our experiments on 
MNIST \cite{lecun1998gradient}, CIFAR-10 \cite{krizhevsky2009learning}, and ImageNet~\cite{DBLP:conf/cvpr/DengDSLL009}. 
For  MNIST and CIFAR-10, we train and verify all three convolutional neural networks (CNNs), i.e., DM-small, DM-medium, and DM-large \cite{DBLP:journals/corr/abs-1810-12715, DBLP:conf/iclr/ZhangCXGSLBH20}, and two fully connected neural networks 
 (FNNs) with three and five layers, respectively. 
The batch size is set 128. We use cross-entropy loss function and Adam~\cite{DBLP:journals/corr/KingmaB14} optimizer to update the parameters. The learning rate decreases following the values of the cosine function between $0$ and $\pi$ after a warmup period during which it increases linearly between $0$ and $1$~\cite{DBLP:conf/emnlp/WolfDSCDMCRLFDS20}. 

For ImageNet, we use the  AlexNet \cite{DBLP:journals/cacm/KrizhevskySH17}, 
 VGG11 \cite{DBLP:journals/corr/SimonyanZ14a}, 
 Inception V1 \cite{DBLP:conf/cvpr/SzegedyLJSRAEVR15}, and 
 ResNet18 \cite{DBLP:conf/cvpr/HeZRS16}
 architectures, which are winners of the image classification competition ILSVRC\footnote{https://image-net.org/challenges/LSVRC/index.php}. We use the same super-parameters as in their original experiments for training these networks. 

\vspace{1ex}
\noindent
\textbf{Metrics.}
We use two metrics in our comparisons: (\romannumeral1) \emph{verified error}, which is the percentage of images
that are not verified to be robust. We   quantify the precision improvement by $(e^{\prime}-e)/e^{\prime}$, with $e$ and $e^{\prime}$ the verified errors of neural networks trained by \textsc{AbsCert} and the competitor, respectively; and  
(\romannumeral2) \emph{time}, which includes training and/or verification on  the same neural network architecture with the same dataset. 
We compute speedup by $t^{\prime}/t$, with $t$ and $t^{\prime}$ the execution time \textsc{AbsCert} and the competitors, respectively.
As the time for different perturbations $\epsilon$ is almost the same, we report the average time (in Table \ref{tab:exp2} and Table \ref{tab:exp4}).

\vspace{2mm}
\noindent
\textbf{Experimental Setup.} 
All experiments on MNIST and CIFAR-10, as well as AlexNet were conducted on a workstation running Ubuntu 18.04 with one NVIDIA GeForce RTX 3090 Ti GPU. All experiments on VGG11, Inception V1 and ResNet18 were conducted on a workstation running Windows11 with one NVIDIA GeForce RTX 3090 Ti GPU.

\subsection{Experimental Results}
\label{subsec:exp-results}

\noindent \textbf{Experiment \uppercase\expandafter{\romannumeral1}: Effectiveness.}
Table \ref{tab:exp1} shows the comparison results of verified errors for DM-small, DM-medium, and DM-large. \textsc{AbsCert} achieves lower verified errors than the competitors. On MNIST, we obtain up to $95.12\%$, $92.65\%$, and $95.64\%$ improvements over LLM, LBP, and AdvIBP on three neural network models (with $\epsilon$ = 0.4), respectively. On CIFAR-10, we achieve a $81.9\%$ improvement to LLM with $\epsilon=16/255$, and all the improvements are above $49\%$. Note that the publicly available code of LBP does not support CIFAR-10. 
 
Another observation is that the improvement increases as $\epsilon$ becomes larger. 
This implies that,  under larger perturbations, the verified errors of the models trained by \textsc{AbsCert} 
increase less slowly than those trained by the competing approaches.
This reflects that the models trained by \textsc{AbsCert} are more robust than those trained by the three competitors. 

As for the accuracy of the trained networks, Table \ref{tab:small} shows that our method achieves
\textit{higher accuracy} than the competitors for \textit{all} datasets and models under the same perturbations.
Moreover, the decrease speed is much less than the one of the networks trained in competitors. Namely,  \textsc{AbsCert} can better resist perturbations, and the models trained by \textsc{AbsCert} have stronger robustness guarantees.

\vspace{1ex}
\noindent \textbf{Experiment \uppercase\expandafter{\romannumeral2}: Efficiency.} Table \ref{tab:exp2} shows the average training and verification time.
\textsc{AbsCert} consumes less training time than the competitors for \emph{all} datasets and models; in particular, compared to LLM, our method achieves up to 26.43x speedup on DM-large of MNIST. Additionally, LBP and AdvIBP can hardly be applied to CIFAR-10 (3200 epochs required), while \textsc{AbsCert} runs smoothly (needs only 30 epochs). This indicates less time (and memory) overhead for \textsc{AbsCert}.

Regarding the verification overhead, \textsc{AbsCert} achieves up to 602.5x speedup and is scalable to large models trained on CIFAR-10. That is mainly because our verification approach treats the networks as black boxes thanks to the abstraction-based training method. 

\begin{table}  
    \centering
    \caption{Verified errors (\%) of the non-ReLU models trained by \textsc{AbsCert}. FC-3 and FC-5 denote FNNs with 3 and 5 hidden layers. DM-s and DM-m refer to DM-small and DM-medium. 
    }
    \vspace{-2mm}
\setlength{\tabcolsep}{6pt}
    \resizebox{0.48\textwidth}{!} 
    {
    \begin{tabular}{c|R{1.1cm}|c|c|c|c|c|c|c}
        \hline
        \multirow{2}{*}{\textbf{D.S.}} &
        \multirow{2}{*}{\textbf{$\epsilon$}\hspace{5mm}} &
        \multicolumn{3}{c|}{\textbf{Sigmoid}} &
        \multicolumn{4}{c}{\textbf{Tanh}} \\
        \cline{3-9}
         & & {FC-3} & 
         {FC-5} & 
         {DM-s} & 
          
         {FC-3} & 
         {FC-5} & 
         {DM-s} & 
         {DM-m} \\
        \hline
        \multirow{4}{*}{\rotatebox{90}{MNIST}}
        & 0.1 & 4.23 & 5.73 & 2.28 
              & 7.95  & 10.50 & 1.56 & 1.04  \\
        & 0.2 & 4.23 & 5.87 & 2.66 
              & 8.06 & 10.50 & 1.65 & 1.14  \\
        & 0.3 & 4.23 & 5.87 & 2.84 
              & 8.29 & 10.50 & 1.75 & 1.14  \\
        & 0.4 & 5.09 & 6.23 & 3.12 
              & 9.39  & 13.16 & 1.97 & 1.24  \\
        \hline
        \multirow{4}{*}{\rotatebox{90}{CIFAR-10}}
        & ${2}/{255}$  & 53.49 & 58.95 & 36.22 
                & 56.62 & 60.07 & 46.69 & 30.02  \\
        & ${16}/{255}$ & 53.49 & 58.95 & 39.59 
                & 56.62 & 61.30 & 48.14 & 31.02  \\
        & ${32}/{255}$ & 54.41 & 60.05 & 41.39 
                & 57.17 & 61.72 & 49.05 & 32.18  \\
       & ${64}/{255}$ & 57.65 & 62.77 & 43.09 
                & 60.99 & 65.73 & 51.22 & 35.62  \\
        \hline
        
    \end{tabular}
    }
    \label{tab:exp3}
    \vspace{-1mm}
\end{table}

\vspace{1ex}

\noindent \textbf{Experiment \uppercase\expandafter{\romannumeral3}: Applicability.}  
We show that our approach is applicable to both CNNs and FNNs with various activation functions, such as Sigmoid and Tanh. 
Table \ref{tab:exp3} shows the verified errors for both types of neural networks trained by our approach. We observe that the 
verified errors of those neural networks trained on MNIST (resp. CIFAR-10) datasets are all below 14\% (resp. 66\%), which are smaller than the benchmark counterparts verified in the work ~\cite{DBLP:journals/pacmpl/SinghGPV19}. 

\begin{table}[]
    \caption{Verified errors (\%) and training time (s/epoch) of the large models trained by \textsc{AbsCert} on ImageNet. 
    }
    \vspace{-2mm}
    \centering
    \resizebox{0.48\textwidth}{!}{
    \begin{tabular}{c|c|c|c|c|c|c|c|c}
         \hline
         \multirow{2}{*}{$\epsilon$} &
         \multicolumn{2}{c}{\textbf{AlexNet}} & 
         \multicolumn{2}{|c}{\textbf{VGG11}} & 
         \multicolumn{2}{|c}{\textbf{Inception V1}} &
         \multicolumn{2}{|c}{\textbf{ResNet18}}\\
         \cline{2-9}
          & Error & Time & Error & Time & Error & Time & Error & Time \\
         \hline
         $2/255$ & 44.96 & \multirow{3}{*}{508.2}
         & 36.29 & \multirow{3}{*}{2530.3}
         & 41.67 &  \multirow{3}{*}{2183.7}
         & 32.15 & \multirow{3}{*}{212.6} \\
        \cline{1-2}\cline{4-4}\cline{6-6}\cline{8-8}
        $4/255$ & 44.96 & 
         & 36.35 & 
         & 41.67 &
         & 32.25 &  \\
        \cline{1-2}\cline{4-4}\cline{6-6}\cline{8-8}
        $8/255$ & 44.97 & 
         & 36.93 & 
         & 43.12 &  
         & 32.86 &  \\
        \hline
    \end{tabular}
    }
    \label{tab:exp4}
\end{table}

\vspace{1ex}
\noindent \textbf{Experiment \uppercase\expandafter{\romannumeral4}: Scalability.} 
We show our method is scalable with respect to training four larger neural network architectures: AlexNet, VGG11, Inception V1, and ResNet18 on ImageNet. Table \ref{tab:exp4} shows the verified errors and training times.
The number of trainable parameters varies from 11 million to 138 million.  
Compared with the reported errors and training times on those representative large models \cite{DBLP:journals/cacm/KrizhevskySH17, DBLP:journals/corr/SimonyanZ14a, DBLP:conf/cvpr/HeZRS16},
our approach achieves competitive performance. 
It is worth mentioning that the reported errors are computed on the testing sets but not verified because those models are too large and cannot be verified by existing verification methods~\cite{DBLP:journals/pacmpl/SinghGPV19, zhang2018efficient, DBLP:conf/cav/KatzBDJK17, DBLP:conf/nips/WangZXLJHK21, DBLP:conf/nips/WangPWYJ18, DBLP:conf/nips/SinghGPV19} due to the high computational complexity. In contrast, \textsc{AbsCert} can verify fairly large networks for its black-box feature.

\vspace{-2mm}
\section{Related Work}\label{sec:rela}
\vspace{-1mm}

This work has been inspired by
earlier  efforts on training and verifying robust models by interval-based abstractions. 

\vspace{1ex}

\noindent
\textbf{Interval Neural Networks (INNs).} A neural network is called an interval neural network if at least one of its input, output, or weight sets are interval-valued \cite{beheshti1998interval}. Interval-valued inputs can capture the uncertainty, inaccuracy, or variability of datasets and thus are used to train prediction models of uncertain systems such as stock markets \cite{roque2007imlp}. Yang and Wu proposed a gradient-based method for smoothing INNs to avoid the oscillation of training \cite{yang2012smoothing}. 
Oala et al. recently proposed to train INNs for image reconstruction with theoretically justified uncertainty scores of predictions   
\cite{oala2020interval}. All these works demonstrate that training on interval-valued data can improve the prediction accuracy under uncertainties. This is consistent with the robustness improvement for image classifications in this work. 

Prabhakar and Afzal proposed to transform regular neural networks to over-approximated INNs for robustness verification \cite{prabhakar2019abstraction}. However, the transformation inevitably introduces overestimation to the models and verification results. Our approach avoids any over-approximation to the trained models by training on interval-valued data. 

\vspace{1ex}
\noindent
\textbf{Verification-in-the-loop Training}. Many approaches on training neural networks with robustness guarantees have been proposed \cite{DBLP:conf/iclr/RaghunathanSL18,DBLP:conf/icml/WongK18,DBLP:conf/nips/WongSMK18,DBLP:conf/icml/MirmanGV18,DBLP:journals/corr/abs-1810-12715,DBLP:conf/iclr/ZhangCXGSLBH20,DBLP:conf/aaai/FanL21}. Most of them are based on linear relaxation \cite{DBLP:conf/icml/WongK18,DBLP:conf/nips/WongSMK18,DBLP:conf/iclr/ZhangCXGSLBH20} or bound propagation \cite{DBLP:conf/icml/MirmanGV18,DBLP:journals/corr/abs-1810-12715,DBLP:conf/iclr/ZhangCXGSLBH20,DBLP:conf/aaai/FanL21}. Linear relaxation-based methods
use linear relaxation
to obtain a convex outer approximation within a norm-bounded perturbation, which results in high time and memory costs \cite{DBLP:conf/icml/WongK18,DBLP:conf/nips/WongSMK18}. In contrast, bound propagation methods are more effective. 
 Gowal et al. \cite{DBLP:journals/corr/abs-1810-12715} proposed IBP to train provably robust neural networks on a relatively large scale. However, the bound it produces can be too loose to be put into practice. Zhang et al.  \cite{DBLP:conf/iclr/ZhangCXGSLBH20} improved IBP by combining the fast IBP bounds in
a forward propagation
and a tight linear relaxation-based bound, CROWN, in a backward propagation. Lee et al. \cite{DBLP:conf/nips/LeeLPL21} also proposed a training method based on linear approximation but considered another important factor - the smoothness of loss function. However, these methods rely heavily on  verification process, so the time complexity is relatively high. AdvIBP~\cite{DBLP:conf/aaai/FanL21} computes the adversarial loss using FGSM and random initialization and computes the robust loss using IBP. However, the verified errors obtained by AdvIBP are relatively higher. Thanks to the abstraction,  networks trained by our method are more robust.

\section{Conclusion}\label{sec:conclu}
\vspace{-1mm}

We have presented a novel, abstraction-based method, naturally coupled with a black-box verification algorithm, for efficiently training provably robust image classifiers.
The experimental results showed that our approach outperformed the state-of-the-art robust training approaches with up to 95.64\% improvement in reducing verified errors. 
Moreover, thanks to its black-box feature, our verification algorithm is more amenable and scalable to  
large neural networks with up to 138 million trainable parameters and is applicable to a wide range of neural networks.

\vspace{2mm}
\noindent
\textbf{ACKNOWLEDGMENTS}
\vspace{1mm}

\noindent
This work was supported by the National Key Research and Development (2019YFA0706404), the National Nature Science Foundation of China (61972150), Huawei Technologies, the NSFC-ISF Joint Program (62161146001), the Fundamental Research Funds for Central Universities, Shanghai Trusted Industry Internet Software Collaborative Innovation Center, and Shanghai International Joint Lab of Trust-worthy Intelligent Software
(22510750100).
 Corresponding authors are Jing Liu and Min Zhang (\texttt{\{jliu, zhangmin\}@sei.ecnu.edu.cn}).

\clearpage
\small
\balance

\clearpage
\nobalance

\noindent
\textbf{A.} \textbf{Calculation of Variance}

\begin{align}
	\label{variance_app}
	D(\mathbb{I}) & = E((\mathbb{I})^2) - E(\mathbb{I})^2 \notag\\
	& = E((\mathbb{I})^2) \notag\\
	& = \underbrace{(-1 + \frac{d}{2})^2 + (-1 + \frac{3d}{2}^2) + \ldots + (1 - \frac{d}{2})^2}_{2/d \enspace items} \notag\\
	& = 2 * \underbrace{((\frac{d}{2})^2 + \ldots + (1 - \frac{d}{2})^2)}_{1/d \enspace items} \\
	& \overset{\text{let n = 1/d}}{=} 2 * \underbrace{((\frac{1}{2n})^2 + \ldots + (\frac{2n-1}{2n})^2)}_{n \enspace items} \notag\\
	& = \frac{1^2 + 3^2 + \ldots + (2n-1)^2}{2n^2} * n \notag\\
	& = \frac{4n^2-1}{6} \notag
\end{align}

\noindent
\textbf{B.} \textbf{Effects of Abstraction Granularity}

\begin{figure}[h!]
    \centering
    \begin{subfigure}{0.49\linewidth}
        \centering
        \includegraphics[width=0.95\linewidth]{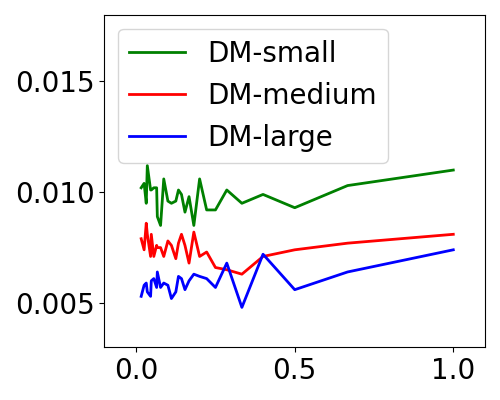}
        \caption{MNIST}
    \end{subfigure}
    \begin{subfigure}{0.49\linewidth}
        \centering
        \includegraphics[width=0.95\linewidth]{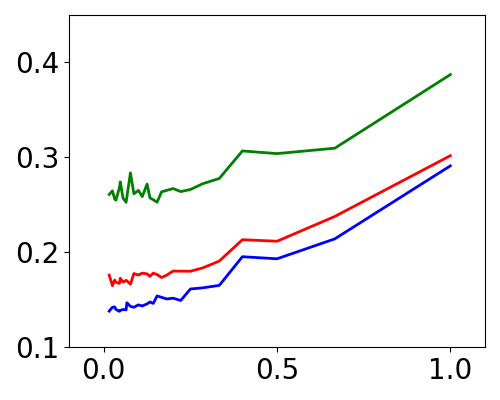}
        \caption{CIFAR-10 and ImageNet}
    \end{subfigure}
    \caption{Measuring verified errors with varying abstraction granularity $d$ ($x$-axis: the size of  granularity; $y$-axis: the verified error). 
    }
    \label{fig:exp3}
\end{figure}

We had investigated the effects of abstraction granularity and obtained some preliminary results. As shown in Figure~\ref{fig:exp3}, 
when the abstraction granularity is relatively small (below the robustness bound), the verified errors are less affected. 
After it exceeds the bound, the verified errors become higher as the abstraction granularity increases. Hence, it is fair to say that  abstraction granularity is a key hyper-parameter for training robust models with low verified errors. We believe that this work would inspire more studies on investigating new mechanisms for finding optimal abstraction granularity.

\end{document}